\documentclass[12pt]{iopart}

\usepackage{algorithm}
\usepackage{algpseudocode}
\usepackage{amsmath,amssymb,amsfonts}
\usepackage{xcolor}
\usepackage{graphicx}
\usepackage{hyperref}
\hypersetup{
  colorlinks   = true, 
  urlcolor     = blue, 
  linkcolor    = blue, 
  citecolor   = blue 
}
\usepackage{caption}
\usepackage[style=authoryear, 
            backend=biber,
            maxcitenames=2,
            minbibnames=3,
            doi=false,
            isbn=false,
            url=true]{biblatex}
\renewbibmacro{in:}{}

\DeclareFieldFormat[article]{volume}{#1}
\DeclareFieldFormat[article]{number}{\mkbibparens{#1}}
\DeclareFieldFormat[article]{pages}{#1}
\renewbibmacro*{volume+number+eid}{%
  \printfield{volume}%
  \printfield{number}%
  \setunit{\addcolon}%
  \printfield{eid}}

\begin{document}

\title[EEG-Reptile]{EEG-Reptile: An Automatized Reptile-Based Meta-Learning Library for BCIs}

\author{Daniil A. Berdyshev$^{1,3}$, Artem M. Grachev$^2$, Sergei L. Shishkin$^1$ and Bogdan L. Kozyrskiy$^4$}

\address{$^1$ MEG Center, Moscow State University of Psychology and Education, Moscow, Russia}
\address{$^2$ Independent researcher, Berlin, Germany}
\address{$^3$ Faculty of Computational Mathematics and Cybernetic, Lomonosov Moscow State University, Moscow, Russia}
\address{$^4$ Independent researcher, Lille, France}
\ead{\mailto{danberdyshev@gmail.com}}
\vspace{10pt}
\begin{indented}
\item[]January 2025
\end{indented}

\begin{abstract}


Meta-learning, i.e., “learning to learn”, is a promising approach to enable efficient BCI classifier training with limited amounts of data.
It can effectively use collections of in some way similar classification tasks, with rapid adaptation to new tasks where only minimal data are available.
However, applying meta-learning to existing classifiers and BCI tasks requires significant effort. To address this issue, we propose EEG-Reptile, an automated library that leverages meta-learning to improve classification accuracy of neural networks in BCIs and other EEG-based applications. It utilizes the Reptile meta-learning algorithm to adapt neural network classifiers of EEG data to the inter-subject domain, allowing for more efficient fine-tuning for a new subject on a small amount of data.
The proposed library incorporates an automated hyperparameter tuning module, a data management pipeline, and an implementation of the Reptile meta-learning algorithm. EEG-Reptile automation level allows using it without deep understanding of meta-learning. We demonstrate the effectiveness of EEG-Reptile on two benchmark datasets (BCI IV 2a, Lee2019 MI) and three neural network architectures (EEGNet, FBCNet, EEG-Inception). Our library achieved improvement in both zero-shot and few-shot learning scenarios compared to traditional transfer learning approaches.

\end{abstract}

%
\section{Introduction}

Brain-Computer Interfaces (BCIs) enable direct communication between the brain and external devices by translating neural activity into commands, bypassing neuromuscular pathways (\cite{ wolpaw2002brain}). By allowing control of prosthetic limbs, exoskeletons, or computer interfaces through neural signals, BCIs can restore movement and autonomy, improving rehabilitation and quality of life. In particular, electroencephalography (EEG) based noninvasive BCIs employing motor imagery (MI) are increasingly used in neurorehabilitation, particularly for post-stroke patients, where they may facilitate motor function recovery via neuroplasticity (\cite{daly2008brain, murphy2009plasticity, frolov2017post, cervera2018brain, mane2020bci, mane2022poststroke}). Users of the MI BCIs imagine specific movements without actual muscle activity, generating neural patterns that can be decoded into control commands.\par

The effectiveness of an EEG based BCI greatly depends on performance of its core computational part, a classifier, which translates patterns of the recorded signals into commands. A BCI classifier needs to be trained on its user’s individual data or at least fine-tuned to it, due to great intersubject variability of the neural data. This variability stems primarily from individual differences in brain anatomy and function. Even higher variability is observed among post-stroke patients, as stroke-induced damage alters neural signals in different ways. Moreover, different electrode positioning, recovering or pathological processes, slow fluctuations of a patient's physiological state and sources of artifacts often make necessary classifier re-calibration even on daily basis. However, the amount of training data which can be obtained from a single user, especially in a single session, is limited. For a range of practical reasons, especially in patients with significant disability, it is highly desirable to make procedures of training data collection as short as possible or even exclude them completely. 

To address these issues, various transfer-learning techniques are being applied to EEG-based BCIs (\cite{azab2018reviewBCI_TF}, \cite{huang2022relation}, \cite{Duan_2023}, \cite{li2024transfer}). The core idea of transfer learning is to find common features between various user sessions and to perform classification based on these features (\cite{WU2022235}). 
However, achieving this requires a large and relatively homogeneous dataset to effectively extract these shared features (\cite{azab2018reviewBCI_TF}, \cite{guetschel2024review}). In addition, the extraction of these features often requires the modification of the original classifier, which prevents the building of more automated solutions (\cite{Cai17092024}).

In contrast, meta-learning focuses on learning how to learn by leveraging a still large dataset, but consisting of smaller, sometimes disparate tasks. This approach enables rapid adaptation to entirely new tasks from minimal data. More broadly, meta-learning (\cite{DBLP:phd/de/Schmidhuber09}, \cite{DBLP:books/sp/1998TP}), also referred to as “learning to learn,” involves training models on a range of tasks or datasets so that they can quickly adapt to novel tasks with minimal additional data.

Consequently, we find a meta-learning approach more suitable for BCI problems. The available pre-training data often comes from numerous distinct BCI users, each with only a small amount of data, and can thus be treated as separate tasks. Instead of searching for common features across all these users, which may not exist at all, it is more promising to train the classifier to quickly adapt to each new user.     

Being more specific, meta-learning approach 
provides an initialization scheme for neural network parameters, that allows to adapt to new BCI users using very small amount of data. This initialization scheme is constructed using data from other users.  
This approach can:

\begin{itemize}
    \item \textit{Enhance Adaptability}: Enable models to generalize across different users and sessions by capturing common patterns in EEG data.
    \item \textit{Reduce Calibration Time}: Minimize the per-user data required for effective model tuning.
    \item \textit{Improve Performance in Variable Conditions}: Handle variability due to physiological differences or neural alterations.
\end{itemize}

However, some patients exhibit neural patterns that significantly deviate from the average due to unique physiological or pathological conditions. Even with high-quality EEG data, their brain activity may not align with common patterns learned during meta-training. Including such outlier data in the meta-learning pool can degrade the model's generalization ability. Therefore, it is crucial to detect and handle these atypical patients separately, ensuring that the model is trained using representative data while developing methods to adapt to individual differences during personalization.\par


Despite previous efforts (\cite{wu_does_2022}, \cite{duan2020meta}) to apply meta-learning algorithms for BCIs, the focus on reducing the size of training datasets has been limited. Specifically, Reptile (\cite{nichol_first-order_2018}) and Model-Agnostic Meta-Learning (MAML) (\cite{finn_model-agnostic_2017}) have not been explored in scenarios where minimal EEG data is used for fine-tuning or zero-shot learning. Notably, previous studies on MAML (\cite{duan2020meta}) involved moderate reductions in dataset size, while other approach lost model agnostic nature (\cite{tremmel2022meta}, \cite{han_meta-eeg_2024}). We propose that the application of general-purpose meta-learning algorithms will enable us to reduce the amount of training data, needed for adaptation of the neural network for a new subject.\par

Recent studies have demonstrated approaches for inter-subject adaptation of neural network classifiers for EEG data. In a recent study (\cite{ng2024subject}), researchers employed extensively redesigned MAML like meta-learning algorithm, achieving promising results in few-shot and zero-shot scenarios. However, this approach involved a more complex meta-learning procedure, which hindered the adoption of new neural network architectures for this method. Furthermore, hyperparameter optimization for the meta-learning algorithm was not automated. 
Meta-learning library presented in our work enables automatic hyperparameter tuning for meta-learning, seamless integration of new neural network architectures. It can be applied with the amount of data available in public datasets, and it is less computationally expensive compared to foundation models and Self-Supervised Learning (SSL) approaches.\par

In this study with our proposed meta-learning library EEG-Reptile\footnote{EEG-Reptile GitHub: \url{https://github.com/gasiki/EEG-Reptile}}, we focus on the Motor Imagery (MI) paradigm to illustrate the advantages of our approach. MI is widely used in real-world BCIs, enabling control of external devices by imagining movements. This is crucial for post-stroke rehabilitation, where MI can activate neural pathways associated with movement, promoting neuroplasticity and aiding recovery. However, MI tasks present significant challenges due to high inter-subject variability and limited per-user data. By focusing on MI data, we aim to demonstrate how meta-learning can effectively handle variability and data scarcity, leading to more adaptable and efficient EEG-based BCIs for rehabilitation purposes.
Our main contributions are:
\begin{itemize}
    \item We propose a method to filter subjects for model pre-training, ensuring a coherent pre-training subset and enhancing subsequent meta-learning performance.
    \item We introduce an advanced meta-learning procedure for EEGNet, a widely used neural network for EEG classification, enabling it to operate effectively in few-shot and even zero-shot training regimes.
    \item We propose an approach for tuning meta-learning hyperparameters on the fly.
    \item We develop a highly automated meta-learning library and demonstrate its effectiveness on several neural network architectures and EEG datasets.
\end{itemize}

\section{Methods}
In our study, we present an approach for meta-learning based on the Reptile algorithm for BCI applications. In this section, we describe used methods, machine learning models and datasets used for training. In addition, we present data preprocessing and algorithm of the proposed meta-learning method.
\subsection{Datasets}
The BCI-IV (2a) (\cite{tangermann_review_2012}) and Lee 2019 (MI) (\cite{lee_eeg_2019}) datasets were used to train ML models and evaluate performance of the proposed library. The datasets were loaded using the MOABB library (\cite{Aristimunha_Mother_of_all_2023}). Data preprocessing was performed utilizing the Braindecode toolbox (\cite{HBM:HBM23730}). Both datasets are recorded using an imaginary movement paradigm. This paradigm is interesting due to the differences in signal between users, which complicates the transfer learning process. It is also possible to use this paradigm to construct BCI systems.
\subsubsection{BCI IV (2a)}
The BCI iv 2a dataset (\cite{tangermann_review_2012}) comprises electroencephalographic (EEG) recordings for 4 motor imagery tasks: right hand, left hand, both feet and tongue. The dataset consists of 22-channel EEG data from 9 subjects, recorded at a sampling rate of 250 Hz. A total of 5184 epochs are included in the dataset, 144 epochs per class for each subject.
Prior to analysis, a band-pass filter (4-38 Hz) and exponential moving standardization were applied to each EEG channel. The duration of each data epoch was 4.5 seconds or 1125 measurements.

\subsubsection{Lee2019 MI}
The Lee2019 MI dataset (\cite{lee_eeg_2019}) represents motor imagery tasks comprising two classes: right hand and left hand. The data consists of 62 EEG channels, recorded at a sampling rate of 1000 Hz. The dataset contains a total of 200 epochs per class to each of 54 subjects included in the dataset.\par
The dataset was preprocessed to ensure compatibility with neural network architectures designed for EEG analysis. Specifically, it was downsampled to 250 Hz to reduce the dimensionality of the data. Then a band-pass filter (4-38 Hz) was applied, followed by exponential moving standardization. Each epoch has a duration of 2.5 seconds and consists of 625 time points. To minimize the dimensionality of the dataset and reduce computational requirements, we selectively retained only 20 EEG channels that have been previously identified as relevant for motor imagery tasks in the work by Lee et al. The selected channels were FC5, FC3, FC1, FC2, FC4, FC6, C5, C3, C1, Cz, C2, C4, C6, CP5, CP3, CP1, CPz, CP2, CP4, CP6.

\subsection{Models}
In this study, we present a meta-learning approach, which can apply meta-learning to any neural network trained with techniques similar to Stochastic Gradient Descent (SGD). To evaluate the efficacy of the proposed approach, we employed three distinct neural networks commonly used for EEG signal classification in MI tasks: EEGNet (\cite{lawhern_eegnet_2018}), EEG-Inception (Motor Imagery) (\cite{zhang_eeg-inception_2021}), and FBCNet (\cite{mane_fbcnet_2021}). These networks were selected for their ability to effectively address the challenges of EEG-based brain-computer interfaces. \par
EEGNet is a compact convolutional neural network designed specifically for EEG-based brain-computer interfaces. Inspired by the work of Vernon J. Lawhern et al., our revised architecture consists of two distinct groups of layers: spatial feature extractors and classifiers. These groups are separated to facilitate independent training and application of different learning rates, allowing for more efficient optimization. EEGNet effectively encapsulates well-known EEG feature extraction concepts for brain-computer interfaces, leveraging depthwise and separable convolutions to achieve high performance across various BCI paradigms, including P300 visual-evoked potentials, error-related negativity responses (ERN), movement-related cortical potentials (MRCP), and sensory motor rhythms (SMR).\par
Two other neural networks used in our study were employed in their original form.
EEG-Inception (MI), a convolutional neural network architecture designed for accurate and robust classification of EEG-based motor imagery (MI). This network was sourced from the Braindecode Python library (\cite{HBM:HBM23730}).
FBCNet, a novel Filter-Bank Convolutional Network proposed in R. Mane et al. study, which was obtained from the TorchEEG python library (\cite{zhang2024torcheeg}).

\subsection{Reptile meta-learning algorithm}
We utilize the Reptile meta-learning algorithm to ensure robustness across different neural network architectures. This algorithm belongs to the class of Model Agnostic Meta-Learning algorithms(\cite{finn_model-agnostic_2017}), which are applicable to any networks trained using gradient descent. The Reptile algorithm (\cite{nichol_first-order_2018}) optimizes the initial weights $\boldsymbol{\theta}$ of a neural network $f(\boldsymbol{\theta})$ over a distribution of tasks $p(\mathcal{T})$. It adjusts the weights to move closer to a point in the weight space that is approximately equidistant from the optimal weights for each task during the current training step. In the context of EEG data, we treat classification problems for individual BCI users as separate tasks. Consequently, during meta-training, Reptile updates the initial weights of the neural network using an averaged difference between the initial weights $\boldsymbol{\theta}$ and user-specific weights $\boldsymbol{\theta}'_i$.\par

Our implementation of Reptile (Alg. \ref{reptile}) introduces notable features:
\begin{itemize}
    \item multiple learning rate coefficients $\boldsymbol{\beta_1 \And \beta_2}$ for distinct groups of layers within the neural network (layers responsible for processing EEG-specific features $\boldsymbol{\theta_1}$ vs. those responsible for final classification $\boldsymbol{\theta_2}$).
    \item flexibility to utilize any optimizer, such as Adam, to update weights during meta-training.
\end{itemize}
These features enable customized optimization strategies for different parts of the network, making model weight updates more efficient during meta-training.\par

\begin{algorithm}[H]
  \caption{Reptile algorithm}\label{reptile} 
  Require: $p(\mathcal{T})$: distribution over tasks, $\alpha$, $\beta$: meta step size hyperparameters, $M$: number of meta-learning epochs \\
  Initialize the initial parameter vector $\boldsymbol{\theta}$ and cleaned $p'(\mathcal{T})$ using the Algorithm \ref{w-init}
  \begin{algorithmic}[1]
        \For{$i = 0$ to $M$} 
        \State Sample batch of tasks $\mathcal{T}_i \sim p'(\mathcal{T})$ with length $N$
        \For{ all $\mathcal{T}_i$}
        \State Sample $K$ data points $\{\textbf{x}^{(k)},\textbf{y}^{(k)}\}$ from $\mathcal{T}_i$
        \State Evaluate $\nabla_{\boldsymbol{\theta}}L_{\mathcal{T}_i}(f_{\boldsymbol{\theta}})$ w.r.t. $K$ data points
        \State $\boldsymbol{\theta}'_i = \boldsymbol{\theta} - \alpha\nabla_{\boldsymbol{\theta}} L_{\mathcal{T}_i}(f_{\boldsymbol{\theta}})$
        \EndFor
        \If{double meta weights $\boldsymbol{\beta_1 \And \beta_2}$}
        \State Update: $\boldsymbol{\theta_1} \leftarrow \boldsymbol{\theta_1} + \frac{\beta_1}{N}\sum\limits_{N}(\boldsymbol{\theta_1}^{'}_i - \boldsymbol{\theta_1})$
        \State Update: $\boldsymbol{\theta_2} \leftarrow \boldsymbol{\theta_2} + \frac{\beta_2}{N}\sum\limits_{N}(\boldsymbol{\theta_2}^{'}_i - \boldsymbol{\theta_2})$
        \Else
        \State Update: $\boldsymbol{\theta} \leftarrow \boldsymbol{\theta} + \frac{\beta}{N}\sum\limits_{N}(\boldsymbol{\theta}^{'}_i - \boldsymbol{\theta})$
        \EndIf
        \EndFor
        \State return $\boldsymbol{\theta}$ or $\boldsymbol{\theta_1} \And \boldsymbol{\theta_2}$
  \end{algorithmic}
\end{algorithm}

In short, the Reptile implementation in our library leverages the agnostic nature of this meta-learning algorithm and introduces adaptability through the use of multiple coefficients and optimizers, enabling effective adaptation across diverse neural network architectures and tasks.

\subsection{EEG-Reptile Library}
\begin{figure}
\centering
\includegraphics[width=0.8\textwidth]{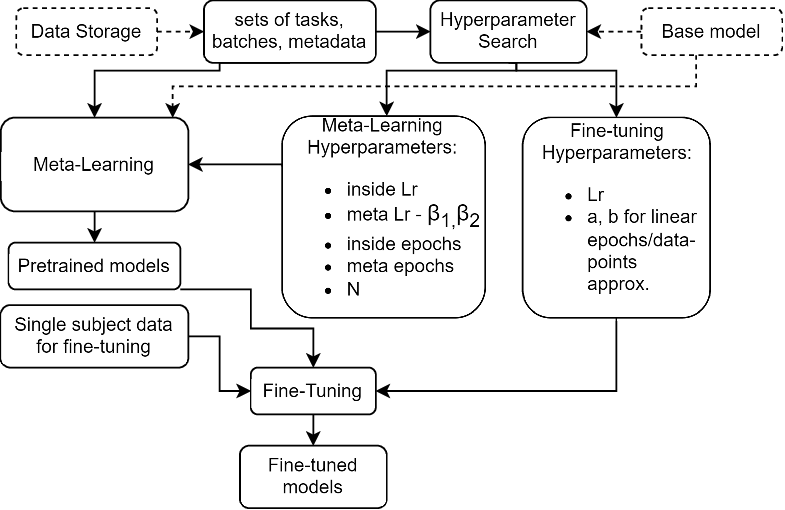}
\caption{Structure of EEG-Reptile library.}\label{Lib_struc}
\end{figure}

We introduce \textbf{EEG-Reptile} 
library, a collection of modules designed for using meta-learning with neural networks for EEG classification, handling EEG datasets necessary for meta-learning, hyperparameter tuning, and model fine-tuning. The library consists of four primary modules: Data Storage, Hyperparameter Search, Meta-Learning, and Fine-Tuning. A simplified scheme of the proposed library is shown on Figure \ref{Lib_struc}.\par
The Data Storage module facilitates the storage and retrieval of preprocessed EEG data from multiple subjects, along with their associated metadata. This module provides prepared datasets for other components within the library.\par
The Hyperparameter Search module leverages the \textbf{Optuna} (\cite{akiba_optuna_2019}) library to perform hyperparameter tuning for both meta-learning and fine-tuning. For meta-learning, this module searches optimal parameters, including:
\begin{itemize}
    \item Number of epochs for meta-learning.
    \item Number of epochs within each meta-step.
    \item Learning rate for training the base model within each meta-step.
    \item Multiple meta-learning rates ($\beta$).
    \item Number of data points used in each meta-step ($N$).
\end{itemize}
For fine-tuning, this module searches optimal parameters, including:
\begin{itemize}
    \item Learning rate.
    \item Linear approximation of the dependence between the number of epochs and the available data points.
\end{itemize}
Linear approximation is necessary because it is impossible to know the size of the fine-tuning set for the target subject in advance and to select a large enough validation set to apply early stopping. In our case, during the selection of hyperparameters on a non-target subject that did not participate in meta-training, we select such linear approximation coefficients $a$ and $b$ so that mean classification quality is maximized for a different size of fine-tuning set.\par
The Meta-Learning module facilitates meta-learning on multiple subjects in various regimes. This module initializes the model weights $\boldsymbol{\theta}$ by computing the average weights $\boldsymbol{\theta}'$ of models trained for a few epochs on each subject. The proposed initialization procedure helps exclude “outlier” subjects whose models have significantly different optimal weight values from the mean of $p(\mathcal{T})$. The proportion of “outliers” removed is denoted by $\gamma$ (Alg. \ref{w-init}).The training regimes vary based on the chosen parameters for the base meta-learning algorithm and the data partitioning strategy. Furthermore, this module supports preparing a network pre-trained on the same datasets using standard Transfer Learning. \par
The Fine-Tuning module enables additional training of a previously meta-trained model on a specific subject. It also gathers statistics on the fine-tuning process and performs testing to assess the fine-tuning performance.
\begin{algorithm}[H]
\caption{Weight Initialization Algorithm}\label{w-init}
\textbf{Input:} $p(\mathcal{T})$: a distribution over $N$ tasks, \quad $\gamma$: outlier removal rate \\
\textbf{Output:} $\boldsymbol{\theta'}$: final weight vector, \quad $p'(\mathcal{T})$

\begin{algorithmic}[1]
\State Initialize $\boldsymbol{\theta}$ randomly
\State Initialize $\boldsymbol{\theta}' = \mathbf{0}$ (same shape as $\boldsymbol{\theta}$)

\For{each task $\mathcal{T}_i$ in $p(\mathcal{T})$}
    \State Train a model from $\boldsymbol{\theta}$ on $\mathcal{T}_i$ to obtain $\boldsymbol{\theta}_i$
    \State $\boldsymbol{\theta}' \gets \boldsymbol{\theta}' + \frac{1}{N}\boldsymbol{\theta}_i$
\EndFor

\For{each $\boldsymbol{\theta}_i$}
    \State $d_i = \left|\, \text{mean}(\boldsymbol{\theta'} - \boldsymbol{\theta}_i) \right|$
\EndFor

\State $n = \lfloor \gamma N \rfloor$ \Comment{Number of outliers to remove}
\State Obtain $p'(\mathcal{T})$ by removing $n$ tasks with the largest $d_i$ from $p(\mathcal{T})$

\State Reset $\boldsymbol{\theta}' = \mathbf{0}$
\For{$\mathcal{T}_i$ in $p'(\mathcal{T})$}
    \State $\boldsymbol{\theta}' \gets \boldsymbol{\theta}' + \frac{1}{(N - n)}\boldsymbol{\theta}_i$
\EndFor

\State \textbf{return} $\boldsymbol{\theta'}$, $p'(\mathcal{T})$
\end{algorithmic}
\end{algorithm}

\subsection{Experimental Setup}
\begin{figure}
\centering
\includegraphics[width=0.8\textwidth]{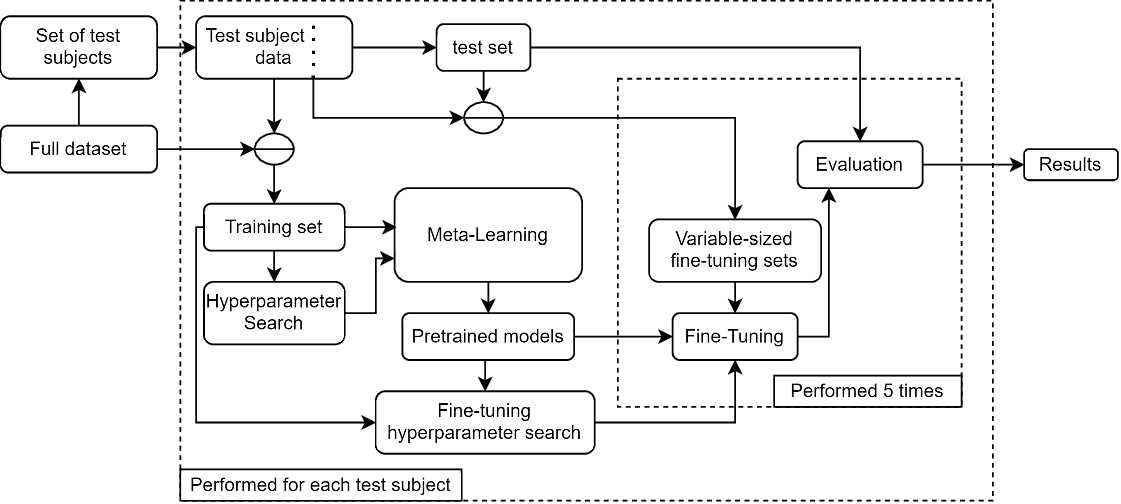}
\caption{Experiment design.}\label{Lib_eval}
\end{figure}

All experiments were conducted using the \textbf{EEG-Reptile} library. Prior to conducting the experiments, we randomly selected five subjects from each dataset and designated them as "test subjects." For each individual experiment, we began by removing data from one test subject in each dataset. We then performed hyperparameter optimization for meta-learning on the remaining subjects' data. After completing the hyperparameter tuning, we carried out meta-training and transfer learning, the latter serving as our baseline approach. Next, we conducted another hyperparameter search to prepare for the fine-tuning stage. Subsequently, we fine-tuned the model on the previously unseen test subject and evaluated its classification accuracy without any additional training (Zero-shot). The full experimental design is illustrated in Figure \ref{Lib_eval}.\par

The model’s performance in each experiment was evaluated using accuracy. This metric was computed with varying amounts of data per class during the fine-tuning stage. Specifically, we measured accuracy as we retrained the model on different numbers of EEG data points (and their corresponding class labels) per class.\par

Each experiment was repeated five times for each test subject, with different random subsets of training data selected in each repetition. For evaluation, we used a fixed test set composed of the last 20\% of each dataset, ensuring equal amounts of data per class. This approach allowed for a fair and consistent performance comparison across all experiments.

\section{Results}

\begin{figure}
\centering
\includegraphics[width=\textwidth]{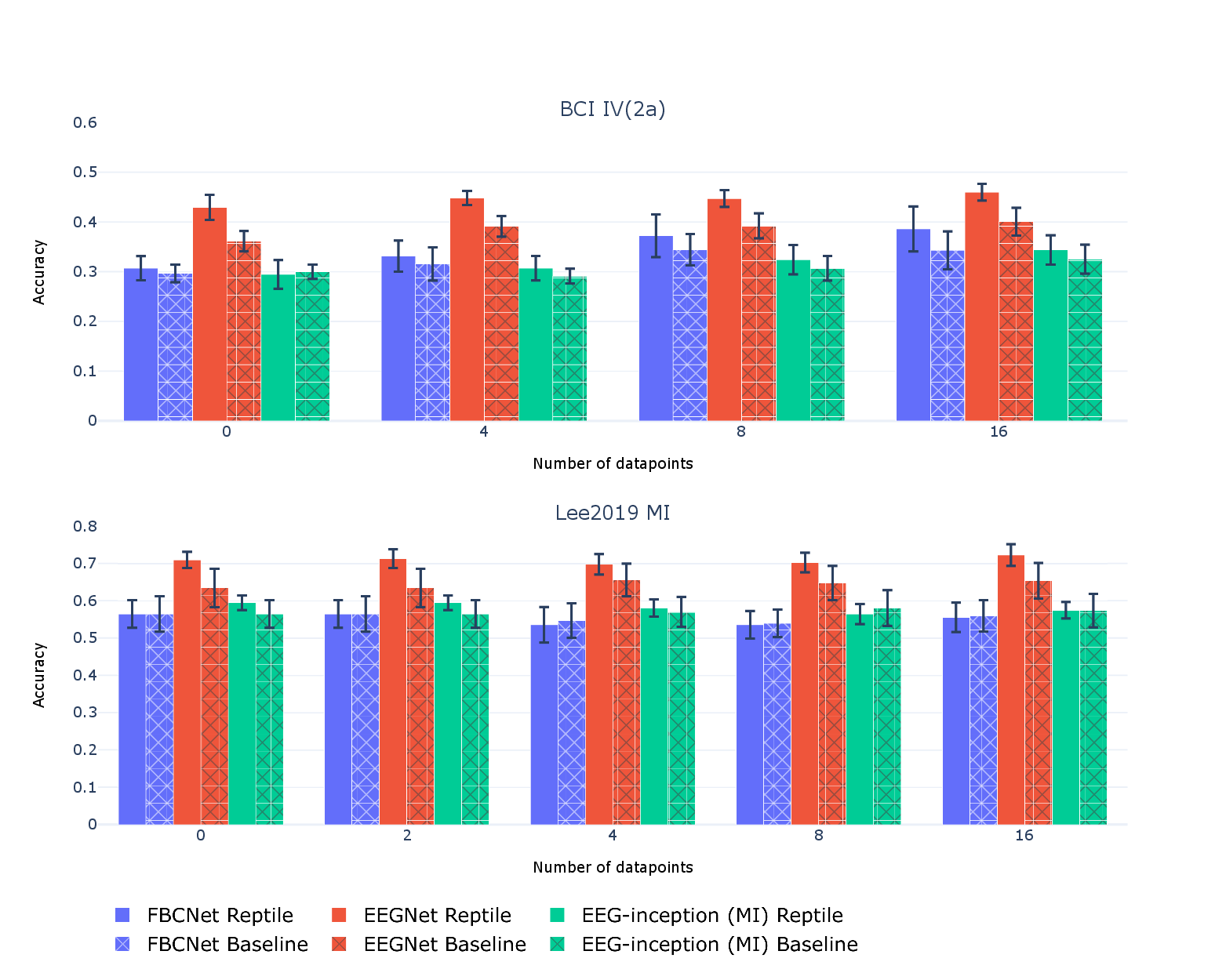}
\caption{Comparison of group average Accuracy with 95\% confidence intervals on BCI IV 2a (four classes) and Lee2019 (MI)  (two classes) datasets for baseline algorithms and algorithms with meta-learning fine-tuned on subsets with different sizes and zero-shot.}
\label{ACC}
\end{figure}

We conducted a comprehensive evaluation of classification performance on the BCI Competition IV 2a dataset (four classes) and the Lee2019 (MI) dataset (two classes). This evaluation was based on the average performance for five randomly selected test subjects from each dataset. The results are presented in Figure \ref{ACC}.

Our experimental design ensured that each test subject remained entirely unseen during the meta-learning and zero-shot testing phases. This approach allowed us to rigorously assess the effectiveness of the machine learning methods without the influence of any fine-tuning steps. \par
For all experiments, fine-tuning was performed on small data subsets of the target subject, with different subset sizes shown on the x-axis. Each individual subset with specified size was randomly chosen five times. Our results demonstrate that meta-learning and transfer learning-based methods exceed performance of random guessing (random guessing would give accuracy of 25\% for 4 class and 50 \% for 2 class), even without fine-tuning on a new subject.\par
Our analysis shows that achieving satisfactory classification quality for EEG data from previously unseen subjects remains challenging in MI classification tasks. This finding highlights the limitations of current state-of-the-art inter-subject transfer learning methods. In particular, we observed that meta-learning approaches still fall short of the standards required for reliable integration into a BCI system, highlighting the need for further research.\par

For the BCI IV 2a dataset, we found that meta-learning with the EEGNet model achieved an average classification accuracy of 43\% $\pm$ 7\% without fine-tuning (zero-shot). Notably, this result was surpassed by fine-tuning on small data subsets, which reached a peak classification accuracy of 46\% $\pm$ 5\% when trained on only 16 data points (4 per class).\par
For the Lee2019 MI dataset, we observed that the highest zero-shot classification accuracy was also achieved using EEGNet with meta-learning, at 71\% $\pm$ 5\%. Furthermore, fine-tuning on small data subsets led to a modest improvement in classification performance, reaching 72\% $\pm$ 7\%, when trained on 16 data points (8 per class).\par

\begin{figure}
\centering
\includegraphics[width=1\textwidth]{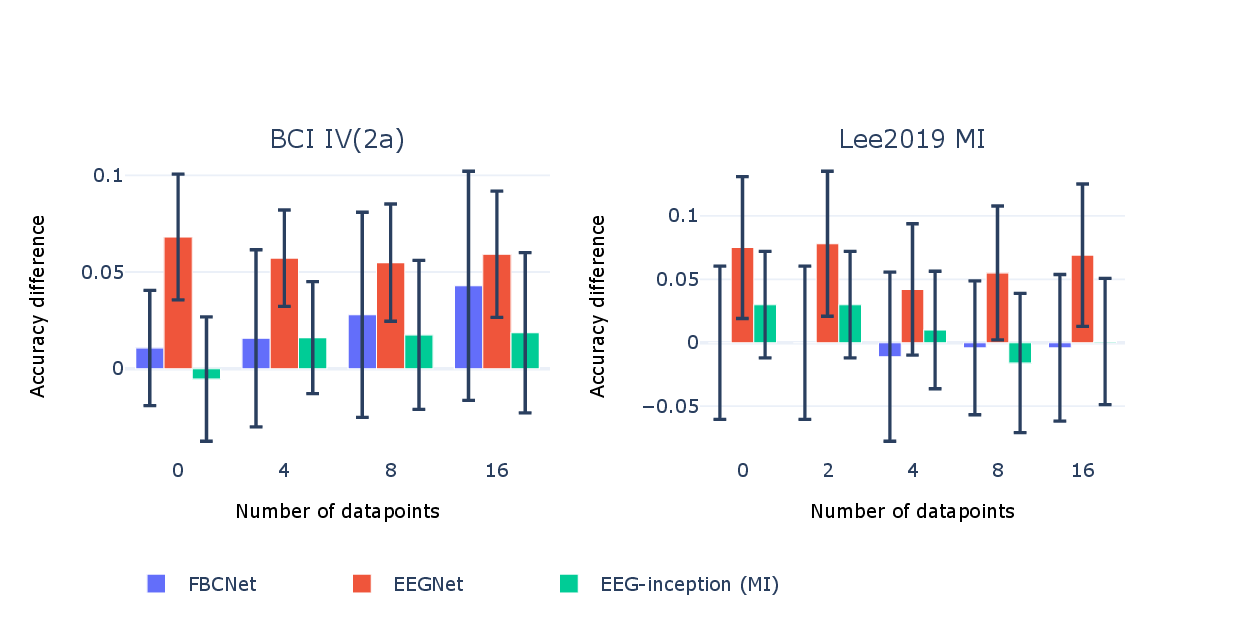}
\caption{Comparison of group average of difference for Accuracy with 95\% confidence intervals on BCI IV 2a (four classes) and Lee2019 (MI) (two classes) datasets between baseline algorithms and algorithms with meta-learning fine-tuned on subsets with different sizes and zero-shot.}\label{diff-ACC}
\end{figure}

We present a graphical comparison (Fig. \ref{diff-ACC}) illustrating the differences in average classification accuracy for each model and dataset under both meta-learning and baseline (transfer learning) conditions. For both datasets, EEGNet with meta-learning significantly outperformed the baseline algorithm (p $<$ 0.05, Wilcoxon signed-rank test). Moreover, the observed improvement in average classification accuracy was statistically significant at the 95\% confidence level.\par

For both datasets, we also observed that when fine-tuned on small data subsets (16 data points), EEGNet with meta-learning again outperformed the baseline algorithm at a significance level of p $<$ 0.05 and a confidence interval of 95\%. This suggests that EEGNet with meta-learning is not only effective for zero-shot classification but also robust when fine-tuned on small data subsets.\par
For other models, we found that two approaches exhibited positive differences in average classification accuracy between meta-learning and baseline algorithms, but these differences were smaller than the confidence interval of 95\%. For the Lee2019 MI dataset, we found that the results for FBCNet and EEG-inception (MI) pre-trained with meta-learning were comparable to those of the baseline algorithm.\par

\begin{figure}
\centering
\includegraphics[width=1\textwidth]{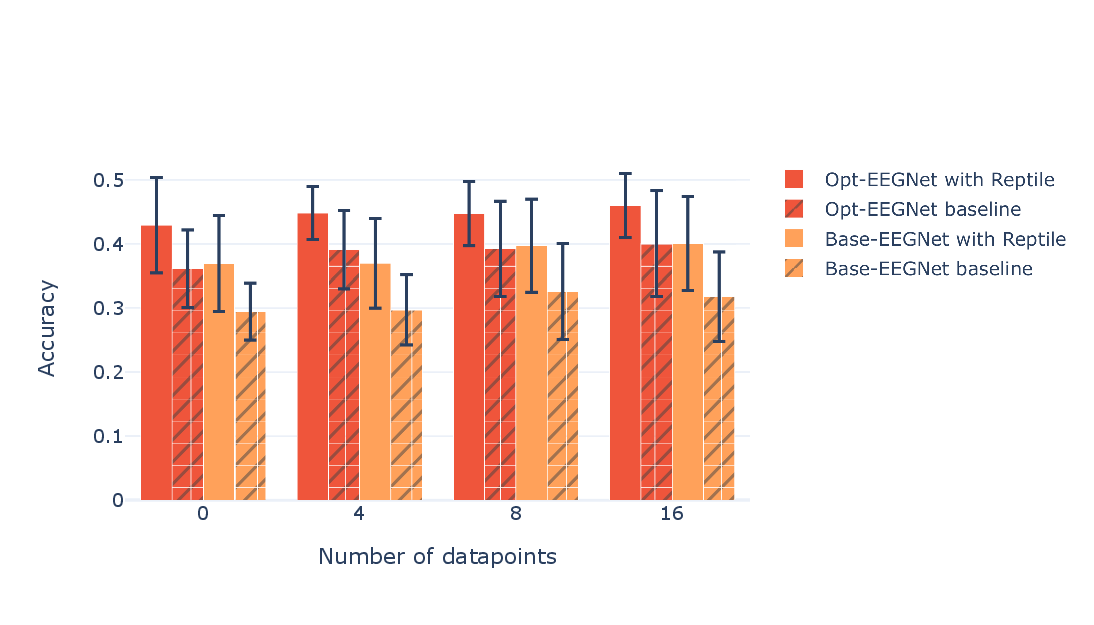}
\caption{Comparison of group average Accuracy with 95\% confidence intervals on BCI IV 2a (four classes) dataset between baseline algorithms and algorithms with meta-learning applied with EEGNet optimization and without, fine-tuned on subsets with different sizes and zero-shot.}\label{mod-ACC}
\end{figure}

Given the significant difference in classification accuracy between the EEG-Net architecture and the other models, we evaluated EEGNet with the proposed optimizations (hereafter referred to as Opt-EEGNet) to assess the effectiveness of these modifications. As described previously, the key distinction between the two architectures is that Opt-EEGNet separates convolutional and fully connected layers into distinct groups. This arrangement enables the use of independent learning rates during fine-tuning, the allocation of individual $\beta$ coefficients for meta-learning, and even the option to freeze weights in one group while fine-tuning the other.\par
To evaluate the effect of this modification, we performed a similar experiment using the unmodified EEGNet (see Fig. \ref{mod-ACC}). The results show that separating the layers into distinct groups improves classification accuracy, even under a straightforward transfer learning approach. Notably, both EEGNet and Opt-EEGNet benefit from meta-learning, indicating that this optimization strategy can enhance the quality of inter-subject knowledge transfer and meta-learning.\par

\section{Discussion}
The results demonstrate that the proposed meta-learning library, EEG-Reptile, can be used to improve classification performance. Furthermore, it allows relatively autonomous operation, making it an attractive solution for various applications.\par
Experiments without fine-tuning and with fine-tuning on small datasets (zero- and few-shot) both showed higher accuracy using EEG-Reptile, particularly when paired with an optimized network.  This improvement holds even for small dataset sizes, suggesting that our library can enhance classification accuracy given a sufficiently large, though not excessive, amount of training data for neural networks.  This is consistent with results demonstrated in work (\cite{berdyshev_meta-optimization_2023}). There, it is shown that such outcomes are achievable using meta-learning. It is possible to achieve similar or higher accuracy improvement by utilizing EEG-Reptile library in such case.\par
Results for optimized and not optimized EEGNet have demonstrated that such optimization could improve classification accuracy in meta-learning and transfer learning. 
The separation of layers into distinct groups may provide a beneficial optimization strategy for other neural architectures as well. Further research is necessary to fully explore the potential benefits of this approach and its applicability to different models and tasks.\par
One of the key features of EEG-Reptile is its ability to filter and select subjects that are too different from the others, enabling automatic exclusion of overly specific subjects from the meta training set. This feature is provided by the weight initialization procedure presented in this work, which is close to the RANSAC method (\cite{fischler_random_1981}). As far as the authors know, such a method has not been applied for this purpose in meta-learning approaches within the context of BCI. This capability can lead to improved mean performance in inter-subject transfer learning.\par

A recent study (\cite{han_meta-eeg_2024}) demonstrated higher classification accuracy compared to our experiments; however, the key difference is that their models were trained on all available data for each user.
Our results, on the other hand, show a significant improvement in classification performance when using a small number of EEG epochs for unseen users. This was achieved through fully automated hyperparameter optimization for meta-learning and fine-tuning. This stands in contrast to previous works (\cite{ahuja2024harnessing}, \cite{wu_does_2022}, \cite{li_novel_2023}), where researchers failed to demonstrate improved classification accuracy on small datasets with cross-subject transfer learning.
Such small datasets, in turn, represent the primary use case for many practical BCI applications.

One of the main challenges associated with meta-learning is the size of available datasets in the sense of number of users. Collecting more even small sessions may improve the performance of our approach. Another downside of the fully automated meta-optimization could be an extensive search for hyperparameters. In our experiments, it took approximately 24 hours with NVIDIA Tesla P100 GPU.



\section{Conclusion}
In this article, we introduce EEG-Reptile, a meta-learning library designed to enhance classification accuracy in BCI tasks with EEG data. Application of the Reptile meta-learning algorithm using this library improved classification performance on two benchmark datasets, BCI IV 2a and Lee2019 MI, in both zero-shot and few-shot learning scenarios in comparison with a straightforward transfer learning approach. EEG-Reptile provides a practical solution to improve the classification accuracy of EEG data, in the cases of an insufficient amount of data for the target subject, which is essential for various applications such as brain-computer interfaces and cognitive research.\par
The proposed library has additional advantages over existing solutions. Firstly, EEG-Reptile allows mostly autonomous operation, making it a valuable tool for researchers and practitioners who may not have extensive expertise in deep learning or meta-learning. Secondly, the proposed weights initialization procedure enables exclusion of “outlier” subjects from the training set of tasks for meta-learning. Lastly, it can be easily used with various neural network architectures, making it a versatile solution for EEG-based applications.

\ack
This research was funded by the Russian Science Foundation, grant 22-19-00528.

\noindent Special thanks to E.I. Chetkin for his contribution in improving the readability of this paper.
\printbibliography

@misc{nichol_first-order_2018,
      title={On First-Order Meta-Learning Algorithms}, 
      author={Alex Nichol and Joshua Achiam and John Schulman},
      year={2018},
      eprint={1803.02999},
      archivePrefix={arXiv},
      primaryClass={cs.LG},
      url={https://arxiv.org/abs/1803.02999}, 
}

@mastersthesis{DBLP:phd/de/Schmidhuber09,
  author       = {J{\"{u}}rgen Schmidhuber},
  title        = {Evolutionary principles in self-referential learning, or on learning
                  how to learn: The meta-meta-. hook},
  school       = {Technical University of Munich, Germany},
  year         = {1987},
  url          = {https://mediatum.ub.tum.de/813180},
  bibsource    = {dblp computer science bibliography, https://dblp.org}
}

@book{DBLP:books/sp/1998TP,
  editor       = {Sebastian Thrun and
                  Lorien Y. Pratt},
  title        = {Learning to Learn},
  publisher    = {Springer},
  year         = {1998},
  url          = {https://doi.org/10.1007/978-1-4615-5529-2},
  doi          = {10.1007/978-1-4615-5529-2},
  isbn         = {978-1-4613-7527-2},
  bibsource    = {dblp computer science bibliography, https://dblp.org}
}

@article{li2024transfer,
  title={Transfer learning in motor imagery brain computer interface: a review},
  author={Li, Mingai and Xu, Dongqin},
  journal={Journal of Shanghai Jiaotong University (Science)},
  volume={29},
  number={1},
  pages={37--59},
  year={2024},
  publisher={Springer}
}

@article{Cai17092024,
  author = {Miao Cai and Jie Hong},
  title = {Joint multi-feature extraction and transfer learning in motor imagery brain computer interface},
  journal = {Computer Methods in Biomechanics and Biomedical Engineering},
  volume = {0},
  number = {0},
  pages = {1--12},
  year = {2024},
  publisher = {Taylor \& Francis},
  doi = {10.1080/10255842.2024.2404541},
  note ={PMID: 39286921},
  URL = {https://doi.org/10.1080/10255842.2024.2404541},
  eprint = {https://doi.org/10.1080/10255842.2024.2404541}
}

@article{ahuja2024harnessing,
  title={Harnessing Few-Shot Learning for EEG signal classification: a survey of state-of-the-art techniques and future directions},
  author={Ahuja, Chirag and Sethia, Divyashikha},
  journal={Frontiers in Human Neuroscience},
  volume={18},
  pages={1421922},
  year={2024},
  publisher={Frontiers Media SA}
}

@inproceedings{finn_model-agnostic_2017,
	title = {Model-Agnostic Meta-Learning for Fast Adaptation of Deep Networks},
	volume = {70},
	url = {https://proceedings.mlr.press/v70/finn17a.html},
	series = {Proceedings of Machine Learning Research},
	pages = {1126--1135},
	booktitle = {Proceedings of the 34th International Conference on Machine Learning},
	publisher = {{PMLR}},
	author = {Finn, Chelsea and Abbeel, Pieter and Levine, Sergey},
	editor = {Precup, Doina and Teh, Yee Whye},
	date = {2017-08-06},
}

@article{zhang_eeg-inception_2021,
	title = {{EEG}-inception: an accurate and robust end-to-end neural network for {EEG}-based motor imagery classification},
	volume = {18},
	issn = {1741-2560, 1741-2552},
	url = {https://iopscience.iop.org/article/10.1088/1741-2552/abed81},
	doi = {10.1088/1741-2552/abed81},
	shorttitle = {{EEG}-inception},
	pages = {046014},
	number = {4},
	journaltitle = {Journal of Neural Engineering},
	shortjournal = {J. Neural Eng.},
	author = {Zhang, Ce and Kim, Young-Keun and Eskandarian, Azim},
	date = {2021-08-01},
}

@article{lawhern_eegnet_2018,
	title = {{EEGNet}: a compact convolutional neural network for {EEG}-based brain–computer interfaces},
	volume = {15},
	issn = {1741-2560, 1741-2552},
	url = {https://iopscience.iop.org/article/10.1088/1741-2552/aace8c},
	doi = {10.1088/1741-2552/aace8c},
	shorttitle = {{EEGNet}},
	pages = {056013},
	number = {5},
	journaltitle = {Journal of Neural Engineering},
	shortjournal = {J. Neural Eng.},
	author = {Lawhern, Vernon J and Solon, Amelia J and Waytowich, Nicholas R and Gordon, Stephen M and Hung, Chou P and Lance, Brent J},
	date = {2018-10-01},
}

@misc{mane_fbcnet_2021,
      title={FBCNet: A Multi-view Convolutional Neural Network for Brain-Computer Interface}, 
      author={Ravikiran Mane and Effie Chew and Karen Chua and Kai Keng Ang and Neethu Robinson and A. P. Vinod and Seong-Whan Lee and Cuntai Guan},
      year={2021},
      eprint={2104.01233},
      archivePrefix={arXiv},
      primaryClass={cs.OH},
      url={https://arxiv.org/abs/2104.01233}, 
}

@article{lee_eeg_2019,
  title={EEG dataset and OpenBMI toolbox for three BCI paradigms: An investigation into BCI illiteracy},
  author={Lee, Min-Ho and Kwon, O-Yeon and Kim, Yong-Jeong and Kim, Hong-Kyung and Lee, Young-Eun and Williamson, John and Fazli, Siamac and Lee, Seong-Whan},
  journal={GigaScience},
  volume={8},
  number={5},
  pages={giz002},
  year={2019},
  publisher={Oxford University Press}
}

@article{tangermann_review_2012,
  title={Review of the BCI competition IV},
  author={Tangermann, Michael and M{\"u}ller, Klaus-Robert and Aertsen, Ad and Birbaumer, Niels and Braun, Christoph and Brunner, Clemens and Leeb, Robert and Mehring, Carsten and Miller, Kai J and M{\"u}ller-Putz, Gernot R and others},
  journal={Frontiers in neuroscience},
  volume={6},
  pages={55},
  year={2012},
  publisher={Frontiers Research Foundation}
}

@inproceedings{akiba_optuna_2019,
	location = {Anchorage {AK} {USA}},
	title = {Optuna: A Next-generation Hyperparameter Optimization Framework},
	isbn = {978-1-4503-6201-6},
	url = {https://dl.acm.org/doi/10.1145/3292500.3330701},
	doi = {10.1145/3292500.3330701},
	shorttitle = {Optuna},
	eventtitle = {{KDD} '19: The 25th {ACM} {SIGKDD} Conference on Knowledge Discovery and Data Mining},
	pages = {2623--2631},
	booktitle = {Proceedings of the 25th {ACM} {SIGKDD} International Conference on Knowledge Discovery \& Data Mining},
	publisher = {{ACM}},
	author = {Akiba, Takuya and Sano, Shotaro and Yanase, Toshihiko and Ohta, Takeru and Koyama, Masanori},
	date = {2019-07-25},
	langid = {english},
}

@article{fischler_random_1981,
	title = {Random sample consensus: a paradigm for model fitting with applications to image analysis and automated cartography},
	volume = {24},
	issn = {0001-0782, 1557-7317},
	url = {https://dl.acm.org/doi/10.1145/358669.358692},
	doi = {10.1145/358669.358692},
	shorttitle = {Random sample consensus},
	pages = {381--395},
	number = {6},
	journaltitle = {Communications of the {ACM}},
	shortjournal = {Commun. {ACM}},
	author = {Fischler, Martin A. and Bolles, Robert C.},
	date = {1981-06},
	langid = {english},
}

@software{Aristimunha_Mother_of_all_2023,
  author = {Aristimunha, Bruno and Carrara, Igor and Guetschel, Pierre and Sedlar, Sara and Rodrigues, Pedro and Sosulski, Jan and Narayanan, Divyesh and Bjareholt, Erik and Quentin, Barthelemy and Schirrmeister, Robin Tibor and Kalunga, Emmanuel and Darmet, Ludovic and Gregoire, Cattan and Abdul Hussain, Ali and Gatti, Ramiro and Goncharenko, Vladislav and Thielen, Jordy and Moreau, Thomas and Roy, Yannick and Jayaram, Vinay and Barachant, Alexandre and Chevallier, Sylvain},
  doi = {10.5281/zenodo.10034223},
  title = {{Mother of all BCI Benchmarks}},
  url = {https://github.com/NeuroTechX/moabb},
  version = {1.0.0},
  year = {2023}
}

@inproceedings{wu_does_2022,
	location = {Glasgow, Scotland, United Kingdom},
	title = {Does Meta-Learning Improve {EEG} Motor Imagery Classification?},
	rights = {https://doi.org/10.15223/policy-029},
	isbn = {978-1-72812-782-8},
	url = {https://ieeexplore.ieee.org/document/9871035/},
	doi = {10.1109/EMBC48229.2022.9871035},
	eventtitle = {2022 44th Annual International Conference of the {IEEE} Engineering in Medicine \& Biology Society ({EMBC})},
	pages = {4048--4051},
	booktitle = {2022 44th Annual International Conference of the {IEEE} Engineering in Medicine \& Biology Society ({EMBC})},
	publisher = {{IEEE}},
	author = {Wu, Xiaoli and Chan, Rosa H. M.},
	date = {2022-07-11},
}

@article{li_novel_2023,
	title = {A novel semi-supervised meta learning method for subject-transfer brain–computer interface},
	volume = {163},
	issn = {08936080},
	url = {https://linkinghub.elsevier.com/retrieve/pii/S0893608023001740},
	doi = {10.1016/j.neunet.2023.03.039},
	pages = {195--204},
	journaltitle = {Neural Networks},
	shortjournal = {Neural Networks},
	author = {Li, Jingcong and Wang, Fei and Huang, Haiyun and Qi, Feifei and Pan, Jiahui},
	date = {2023-06},
	langid = {english},
}

@article{han_meta-eeg_2024,
  title={Meta-eeg: Meta-learning-based class-relevant eeg representation learning for zero-calibration brain--computer interfaces},
  author={Han, Ji-Wung and Bak, Soyeon and Kim, Jun-Mo and Choi, WooHyeok and Shin, Dong-Hee and Son, Young-Han and Kam, Tae-Eui},
  journal={Expert Systems with Applications},
  volume={238},
  pages={121986},
  year={2024},
  publisher={Elsevier}
}

@inproceedings{berdyshev_meta-optimization_2023,
	location = {Novosibirsk, Russian Federation},
	title = {Meta-Optimization of Initial Weights for More Effective Few- and Zero-Shot Learning in {BCI} Classification},
	rights = {https://doi.org/10.15223/policy-029},
	isbn = {9798350307979},
	url = {https://ieeexplore.ieee.org/document/10329624/},
	doi = {10.1109/CSGB60362.2023.10329624},
	eventtitle = {2023 {IEEE} Ural-Siberian Conference on Computational Technologies in Cognitive Science, Genomics and Biomedicine ({CSGB})},
	pages = {263--267},
	booktitle = {2023 {IEEE} Ural-Siberian Conference on Computational Technologies in Cognitive Science, Genomics and Biomedicine ({CSGB})},
	publisher = {{IEEE}},
	author = {Berdyshev, Daniil A. and Grachev, Artem M. and Shishkin, Sergei L. and Kozyrskiy, Bogdan L.},
	date = {2023-09-28},
}

@article{HBM:HBM23730,
  title={Deep learning with convolutional neural networks for EEG decoding and visualization},
  author={Schirrmeister, Robin Tibor and Springenberg, Jost Tobias and Fiederer, Lukas Dominique Josef and Glasstetter, Martin and Eggensperger, Katharina and Tangermann, Michael and Hutter, Frank and Burgard, Wolfram and Ball, Tonio},
  journal={Human brain mapping},
  volume={38},
  number={11},
  pages={5391--5420},
  year={2017},
  publisher={Wiley Online Library}
}

@article{zhang2024torcheeg,
  title={TorchEEGEMO: A deep learning toolbox towards EEG-based emotion recognition},
  author={Zhang, Zhi and Zhong, Sheng-hua and Liu, Yan},
  journal={Expert Systems with Applications},
  volume={249},
  pages={123550},
  year={2024},
  publisher={Elsevier}
}

@article{wolpaw2002brain,
  title={Brain--computer interfaces for communication and control},
  author={Wolpaw, Jonathan R and Birbaumer, Niels and McFarland, Dennis J and Pfurtscheller, Gert and Vaughan, Theresa M},
  journal={Clinical neurophysiology},
  volume={113},
  number={6},
  pages={767--791},
  year={2002},
  publisher={Elsevier}
}

@article{daly2008brain,
  title={Brain--computer interfaces in neurological rehabilitation},
  author={Daly, Janis J and Wolpaw, Jonathan R},
  journal={The Lancet Neurology},
  volume={7},
  number={11},
  pages={1032--1043},
  year={2008},
  publisher={Elsevier}
}

@article{frolov2017post,
  title={Post-stroke rehabilitation training with a motor-imagery-based brain-computer interface (BCI)-controlled hand exoskeleton: a randomized controlled multicenter trial},
  author={Frolov, Alexander A and Mokienko, Olesya and Lyukmanov, Roman and Biryukova, Elena and Kotov, Sergey and Turbina, Lydia and Nadareyshvily, Georgy and Bushkova, Yulia},
  journal={Frontiers in neuroscience},
  volume={11},
  pages={400},
  year={2017},
  publisher={Frontiers Media SA}
}

@article{murphy2009plasticity,
  title={Plasticity during stroke recovery: from synapse to behaviour},
  author={Murphy, Timothy H and Corbett, Dale},
  journal={Nature reviews neuroscience},
  volume={10},
  number={12},
  pages={861--872},
  year={2009},
  publisher={Nature Publishing Group UK London}
}

@article{azab2018reviewBCI_TF,
  title={A review on transfer learning approaches in brain--computer interface},
  author={Azab, Ahmed M and Toth, Jake and Mihaylova, Lyudmila S and Arvaneh, Mahnaz},
  journal={Signal processing and machine learning for brain-machine interfaces},
  pages={81--98},
  year={2018}
}

@ARTICLE{duan2020meta,
  author={Duan, Tiehang and Shaikh, Mohammad Abuzar and Chauhan, Mihir and Chu, Jun and Srihari, Rohini K. and Pathak, Archita and Srihari, Sargur N.},
  journal={IEEE Access}, 
  title={Meta Learn on Constrained Transfer Learning for Low Resource Cross Subject EEG Classification}, 
  year={2020},
  volume={8},
  pages={224791-224802},
  doi={10.1109/ACCESS.2020.3045225},
  publisher={IEEE}}

@article{WU2022235,
  title = {Transfer learning for motor imagery based brain–computer interfaces: A tutorial},
  journal = {Neural Networks},
  volume = {153},
  pages = {235-253},
  year = {2022},
  issn = {0893-6080},
  doi = {https://doi.org/10.1016/j.neunet.2022.06.008},
  url = {https://www.sciencedirect.com/science/article/pii/S0893608022002131},
  author = {Dongrui Wu and Xue Jiang and Ruimin Peng},
}

@article{huang2022relation,
  title={Relation Learning Using Temporal Episodes for Motor Imagery Brain-Computer Interfaces},
  author={Huang, Xiuyu and Liang, Shuang and Zhang, Yuanpeng and Zhou, Nan and Pedrycz, Witold and Choi, Kup-Sze},
  journal={IEEE Transactions on Neural Systems and Rehabilitation Engineering},
  volume={31},
  pages={530--543},
  year={2022},
  publisher={IEEE}
}

@article{Duan_2023,
  doi = {10.1088/1741-2552/acaee7},
  url = {https://dx.doi.org/10.1088/1741-2552/acaee7},
  year = {2023},
  month = {1},
  publisher = {IOP Publishing},
  volume = {20},
  number = {1},
  pages = {016026},
  author = {Xu Duan and Songyun Xie and Yanxia Lv and Xinzhou Xie and Klaus Obermayer and Hao Yan},
  title = {A transfer learning-based feedback training motivates the performance of SMR-BCI},
  journal = {Journal of Neural Engineering},
}

@article{cervera2018brain,
  title={Brain-computer interfaces for post-stroke motor rehabilitation: a meta-analysis},
  author={Cervera, Mar{\'\i}a A and Soekadar, Surjo R and Ushiba, Junichi and Mill{\'a}n, Jos{\'e} del R and Liu, Meigen and Birbaumer, Niels and Garipelli, Gangadhar},
  journal={Annals of clinical and translational neurology},
  volume={5},
  number={5},
  pages={651--663},
  year={2018},
  publisher={Wiley Online Library}
}

@article{tremmel2022meta,
  title={A meta-learning BCI for estimating decision confidence},
  author={Tremmel, Christoph and Fernandez-Vargas, Jacobo and Stamos, Dimitris and Cinel, Caterina and Pontil, Massimiliano and Citi, Luca and Poli, Riccardo},
  journal={Journal of Neural Engineering},
  volume={19},
  number={4},
  pages={046009},
  year={2022},
  publisher={IOP Publishing}
}

@article{mane2020bci,
  title={BCI for stroke rehabilitation: motor and beyond},
  author={Mane, Ravikiran and Chouhan, Tushar and Guan, Cuntai},
  journal={Journal of neural engineering},
  volume={17},
  number={4},
  pages={041001},
  year={2020},
  publisher={IOP Publishing}
}

@article{mane2022poststroke,
  title={Poststroke motor, cognitive and speech rehabilitation with brain--computer interface: a perspective review},
  author={Mane, Ravikiran and Wu, Zhenzhou and Wang, David},
  journal={Stroke and vascular neurology},
  volume={7},
  number={6},
  year={2022},
  publisher={BMJ Specialist Journals}
}

@article{guetschel2024review,
  title={Review of deep representation learning techniques for brain--computer interfaces},
  author={Guetschel, Pierre and Ahmadi, Sara and Tangermann, Michael},
  journal={Journal of Neural Engineering},
  volume={21},
  number={6},
  pages={061002},
  year={2024},
  publisher={IOP Publishing}
}

@article{ng2024subject,
  title={Subject-independent meta-learning framework towards optimal training of eeg-based classifiers},
  author={Ng, Han Wei and Guan, Cuntai},
  journal={Neural Networks},
  volume={172},
  pages={106108},
  year={2024},
  publisher={Elsevier}
}
\end{document}